%% file: KDD_2022_more.tex
\documentclass[11pt,3p, final]{elsarticle}
\makeatletter
\def\ps@pprintTitle{%
 \let\@oddhead\@empty
 \let\@evenhead\@empty
 \def\@oddfoot{\centerline{\thepage}}%
 \let\@evenfoot\@oddfoot}
\makeatother

\usepackage[utf8]{inputenc} 
\usepackage[T1]{fontenc}    
\usepackage{hyperref}       
\usepackage{url}            
\usepackage{booktabs}       
\usepackage{amsfonts}       
\usepackage{nicefrac}       
\usepackage{microtype}      
\usepackage{graphicx}
\usepackage{amsmath}
\usepackage{appendix}

\begin{document}

\title{Identifying and Overcoming Transformation Bias in Forecasting Models}

\author{Sushant More \\
morsusha@amazon.com}
\address{Amazon, Seattle,  WA}

\begin{abstract}
\hspace*{0.5cm}
 Log and square root transformations of target variable are routinely used in forecasting models to predict future sales. These transformations often lead to better performing models. However, they also introduce a systematic negative bias (under-forecasting). In this paper, we demonstrate the existence of this bias, dive deep into its root cause and introduce two methods to correct for the bias. We conclude that the proposed bias correction methods improve model performance (by up to $50\%$) and make a case for incorporating bias correction in modeling workflow. 
   
We also experiment with `Tweedie' family of cost functions which circumvents the transformation bias issue by modeling directly on sales. We conclude that Tweedie regression gives the best performance so far when modeling on sales making it a strong alternative to working with a transformed target variable.
\end{abstract}

\maketitle

\section{Introduction}

One of the science goals in Amazon Devices is to determine which device each customer wants to buy, and ensure that every customer’s choice is available for them to purchase at the time and place they desire. To deliver on this goal, the science team has invested in developing device demand forecasting models, which we call Versioned Demand Planning (VDP) models.  The performance of VDP models is usually monitored at a region and device type level.\footnote{Examples of device types are Alexa Echo,  Kindle/ eReader), Tablet,  Fire TV etc.}. 

An investigation of these VDP models indicated that many of the models (built using $log(sales)$ as target variable) suffered from an issue of negative bias or under-forecasting.
\begin{figure}
  \includegraphics[width=0.9\linewidth]{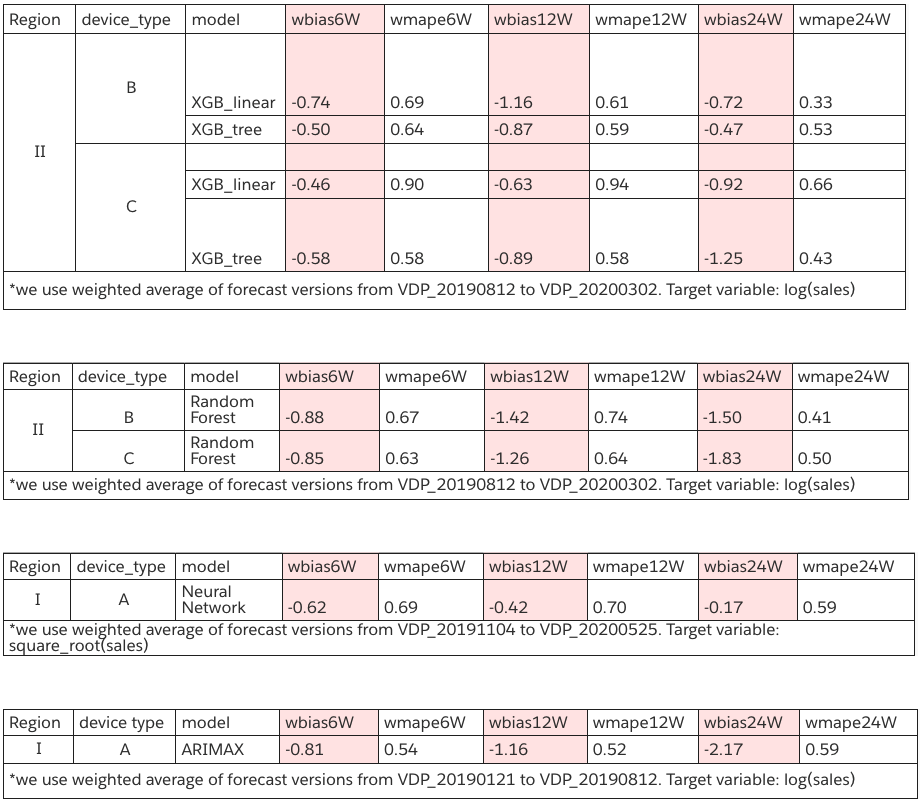}
  \caption{Negative Bias across various models when modeled on a transformed target variable.  Persistent negative bias is seen irrespective of the model type (XGBoost, Random Forest, Linear Regression, Neural Networks, and ARIMAX) and region/ device type combination.  We report relative metrics (Eq.~\eqref{eq:reported_metrics}) for 4 different time horizons (6, 12,  and 24 weeks). }
  \label{fig:various_ models_neg_bias}
\end{figure}
This is demonstrated in Fig.~\ref{fig:various_ models_neg_bias}. 
To maintain business confidentiality,  we anonymize the regions by roman numerals (I,  II,  III,  $\ldots$) and device types by upper-case English letters (A,  B,  C).  Furthermore the metrics -- WMAPE (weighted mean-absolute percentage error) and WBias (weighted bias) are reported relative to the performance of Linear Regression (LR) model for region I and device type A throughout the paper. 
For a given model for a region r, device d,  and time horizon h:
\begin{eqnarray}
\rm reported~WMAPE_{\,r, \,d,\,h, \,model} = \frac{{{WMAPE_{\,r, \,d,\,h,\, model}} }}{WMAPE_{\,r=I, \, d=A,\, h, \,model=LR}}  \nonumber\\
\rm
reported~WBias_{\,r,\, d,\,h, \,model} = \frac{{{WBias_{\, r,\, d,\,h,\,model}} }}{|WBias_{\,r=I, \, d=A, \,h,\,model=LR}|} 
\label{eq:reported_metrics} 
\end{eqnarray}
Please refer to Appendix~\ref{sec:model_metrics} (Eq.~\eqref{eq:mape_bias_def} regarding how we define the MAPE and Bias metrics.  As seen in Fig.~\ref{fig:various_ models_neg_bias},  a systematic negative bias is present regardless of the model choice (XGBoost, Random Forest, Linear Regression, Neural networks, ARIMAX) and is seen irrespective of the region/device type combination. 

The above observation will form the basis of this paper. It is arranged as follows. In the next section, we dive deep into the scientific basis of the observed under-forecasting.  After establishing the root cause for bias in Sec.~\ref{sec:root_cause}, we look into various methods of bias correction in Sec.~\ref{sec:bias_corr_techniques}.  In Sec.~\ref{sec:circumvent_tb}, we look at the methods that circumvent the transformation bias altogether by building models directly on sales without invoking any transformation.  We summarize our findings and give glimpse of the ongoing work in Sec.~\ref{sec:conclusion}. 

Note that even though we use device forecasting as our test ground, the problem of transformation bias that we tackle and the methods we develop are applicable to any regression setting where the target variable is transformed before fitting a Machine Learning model.

\section{Root cause of negative bias}\label{sec:root_cause}

Our goal is to predict daily sales for each individual item given the relevant features such as selling price, seasonality, holidays, in-stock status, and product attributes. We build models on past 2-3 years of sales history and predict daily sales at individual item grain for 6-12 months in future.

Historically, we  found it useful (see Sec.~\ref{subsec:model_on_sales}) to build machine learning (ML) models with $log(sales)$ as target variable and exponentiate the model prediction. To root cause the chronic issue of negative backtesting bias (refer Fig.~\ref{fig:various_ models_neg_bias}), we focus on the simplest ML problem of in-sample model fit by comparing model prediction ($\hat{Y}$) with target variable ($Y$). 

\begin{figure}
\centering
	\includegraphics[width=0.9\linewidth]{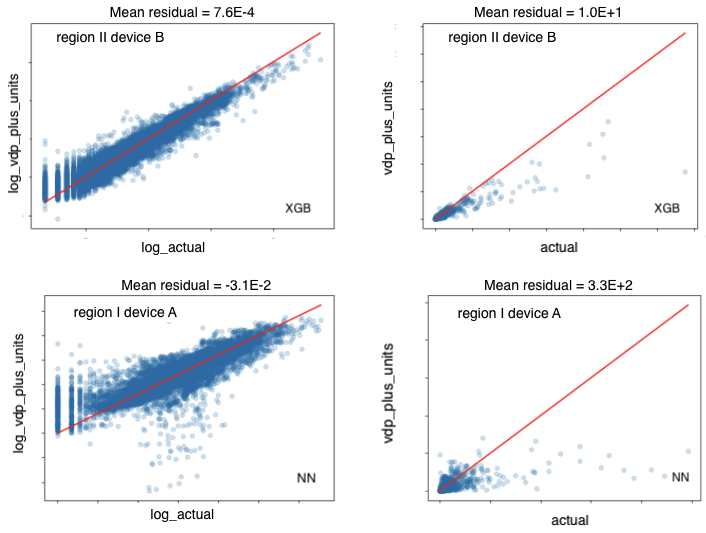}
	\caption{In-sample model fit for two machine learning models for two different device types. The plots on the left are the fit in the modeling units. The X and Y axes are target variable ($Y$) and the prediction output from the ML model ($\hat{Y}$) respectively. The plots on the right are obtained by back-transforming the values from plots on the left. The red line is y = x line. Points below the red line are under-forecasts.}
	\label{fig:in_sample_model_fit}
\end{figure}
In Fig.~\ref{fig:in_sample_model_fit}, the graphs on the left are in-sample model fits in the modeling (i.e, logarithmic) units. Note that we refer to model prediction as ‘vdp\_plus\_units’. The plots on the right is when the same model output is back-transformed (exponentiated) to the original units (without any new model fit). 
${\rm Residual} = Y - \hat{Y}$ 
is the difference between the original value and the prediction. Ideally, we want the residuals to have a mean of zero. Large positive mean for residuals implies a negative bias (or under-forecasting). Fig.~\ref{fig:in_sample_model_fit} shows that:
1)
Models do not show a bias in the modeling units. The mean of residuals is close to zero (refer plots' title). 
2) 
A large negative bias is introduced when the prediction is back-transformed to original units. This is seen by a large positive mean of residuals (plots on right in Fig.~\ref{fig:in_sample_model_fit}). 

Even though, we only demonstrate the effect of target transformation in XGBoost (XGB) and Neural Network (NN) models in Fig.~\ref{fig:in_sample_model_fit}, the behavior is qualitatively similar for other model types as well. Also, it is similar for both logarithm or square-root transformation of the target variable.

So far, we provided an empirical evidence of how modeling on a transformed target variable and back-transforming leads to bias. In the next subsection, we will dive deep into the mathematical basis of this transformation bias.

\subsection{Mathematical basis of transformation bias} \label{subsec:Jensen_inequality}

Jensen's inequality states that the convex transformation of mean is less than or equal to the mean applied after convex transformation \cite{Jensen}. The opposite is true of concave transformations. In context of probability theory, for a random variable $X$ and a concave function $f$: $E[f(X)] <= f(E[X])$. In our case, $X$ is $sales$ and $f$ is typically the logarithm which translates to
\begin{equation}
\label{eq:JI_log}
E(log(sales)) \leq log(E(sales))
\end{equation}
The expectation value, $E$, in Eq.~\eqref{eq:JI_log} is the mean value of forecast distribution for given item on a given day. 
$E(log(sales))$ is what we obtain as output from ML model built on a log transformed target variable and optimized on mean-squared error. Typically, the business is interested in $E(sales)$. To obtain $E(sales)$, we exponentiate the model output which is $E(log(sales))$. 
From Eq.~\eqref{eq:JI_log}, we obtain
\begin{equation}
\label{eq:JI_back_transformed}
exp\left({E(log(sales))}\right) \leq E(sales)\;.
\end{equation}  

Eq.~\eqref{eq:JI_back_transformed} is the mathematical demonstration of our claim that working with a  transformed target variable and back-transforming leads to \emph{Transformation Bias}. Note that even though, we worked with the special case of logarithm transformation, our conclusions hold for any concave transformation. 

Note that in general transformation bias could be one of the several sources of bias (e.g., inaccurate trend, selling price etc.) in the model. The different sources of bias could add up in such a way that no net bias is observed in the back-transformed units. Nonetheless, as demonstrated in Eqs.~\eqref{eq:JI_log} and \eqref{eq:JI_back_transformed}, transformation bias will always be present when working with transformed target variable.

\subsection{Modeling directly on sales}
\label{subsec:model_on_sales}
\begin{figure}
\centering
	\includegraphics[width=0.9\linewidth]{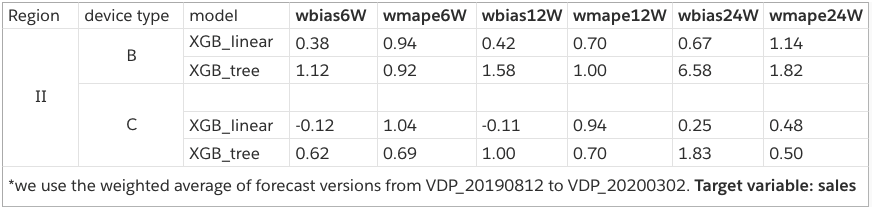}
	\caption{Performance metrics when a forecasting model is built directly on sales as a target variable. 
  XGB\_linear and XGB\_tree are XGBoost models using linear and tree base learners respectively. We are typically interested in how the model performs 6, 12, and 24 weeks in future. The horizon length is based on business requirements.}
	\label{fig:sales_metric}
\end{figure}

One of the straightforward ways to avoid transformation bias is to model directly on sales. Empirically, we find that the performance deteriorates when a model is built directly on sales (Fig.~\ref{fig:sales_metric}). 

\begin{figure}
\centering
	\includegraphics[width=0.9\linewidth]{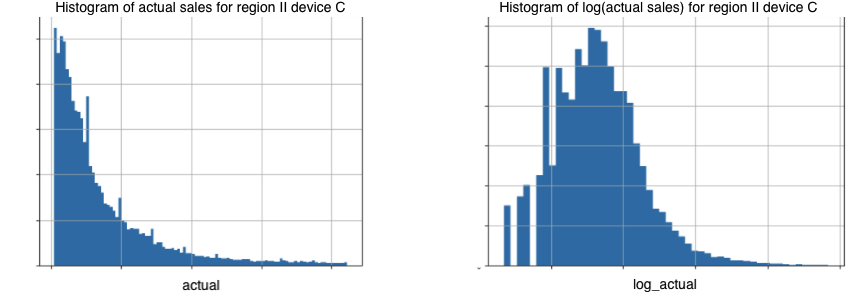}
	\caption{Histogram of daily individual item-level sales data (left) and distribution of data after log transformation (right). In the left plot, we exclude data points over $95^{\rm th}$ quantile for plotability.}
	\label{fig:sales_log_sales_distribution}
\end{figure}

One of the reasons for the poor performance on untransformed target variable is the skewed nature of sales data as demonstrated in Fig.~\ref{fig:sales_log_sales_distribution}. 
The business reason for the right-skewed sales distribution is the spiky sales seasonality. We have days such as Black Friday, Prime Day, and Christmas Holidays when we observe significantly higher sales than most days. Moreover, some items are more popular than others adding to the skewness of distribution. 

This skewed distribution makes it difficult to preserve homoscedasticity and normality of residuals. Additionally, minimizing mean squared-error (MSE) is the maximum likelihood estimator when the noise term is Gaussian. This assumption of a Gaussian noise is typically ill-suited in presence of a highly skewed distribution (left plot in Fig.~\ref{fig:sales_log_sales_distribution}). 

On the other hand, working with $log(sales)$ makes the distribution look more normal (right plot in Fig.~\ref{fig:sales_log_sales_distribution}) and makes the data more amenable to modeling. Logarithm also allows us to better treat the outliers. Note that in our case,  outliers (e.g., Black Friday) are of prime business interest and can't just be discarded. Another advantage of using logarithm is that because of exponentiation during back-transformation, the prediction is always positive which makes sense for forecasting of sales.

We will return to modeling on sales again in Sec.~\ref{sec:circumvent_tb}, but our conclusion based on the discussion so far is that 1) Modeling directly on sales leads to poor performance 2) Modeling on $log(sales)$ has obvious advantages, but introduces a systematic transformation bias. One of the way ahead is to keep using the log transformation, but explicitly correct for the bias introduced. This will be the basis of our discussion in the next section.   

\section{Bias correction techniques}\label{sec:bias_corr_techniques}

Modeling with a transformed variable was acknowledged as problematic since the 1930s \cite{first_paper}. Formal bias correction for logarithm was computed in 1941 \cite{Finney_1941}. In 1960, a general bias correction procedure for all transformations was developed \cite{Neyman_Scott_1960}. In the next two subsections, we will look at two different methods of bias correction. The first method (Sec.~\ref{subsec:weight_addition}) is author's original work and the second (Sec.~\ref{subsec:PBBC}) is an improvement over an existing method in literature \cite{Forestry_paper, newman}.

 \subsection{Addition of weights in cost function}
 \label{subsec:weight_addition}
 
 The standard cost function in regression analysis is the mean-squared error. 
 \begin{equation}
 J = \sum_{i \in \rm samples} w_i (y_i - f(X_i))^2
 \label{eq:weighted_mse}
 \end{equation}
 In the default case, $w_i = 1$ in Eq.~\eqref{eq:weighted_mse}. That is all the points are equally weighted. We find that making $w_i$ in Eq.~\eqref{eq:weighted_mse} a function of sales helps mitigate the negative bias. Our prescription is as follows:
 \begin{enumerate}
 \item
 Start with the default uniform weight ($w_i = 1$) and 
 successively increase the weighting, making it more aggressive (from $\rm log\_sales$ weighting to $\rm sqrt\_sales$ to $\rm sales$ and so on).
 \item
 With successive increases bias increases (from large negative towards zero) and MAPE (mean-absolute-percentage-error) decreases.
 At a certain point bias gets too large and MAPE deteriorates. Stop when the lowest MAPE and bias numbers are obtained. 
 \end{enumerate}
 
 \begin{figure}
\centering
	\includegraphics[width=0.9\linewidth]{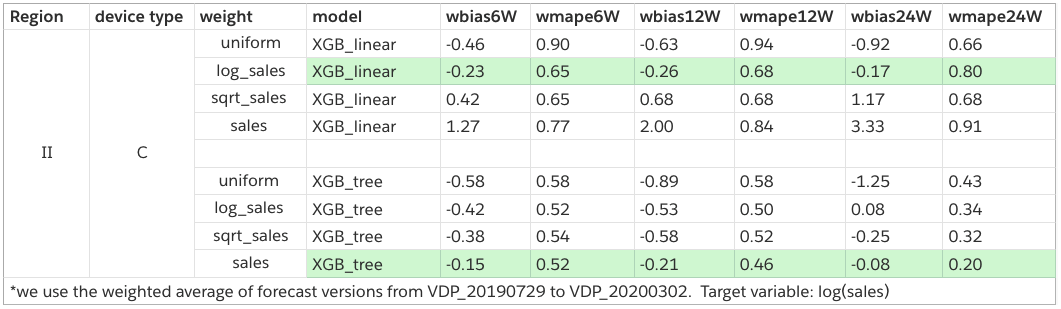}
	\caption{Effect of sales weighting on model performance across different time horizons (6, 12, and 24 weeks).}
	\label{fig:sales_weighting_demo}
\end{figure} 
The effect of successive sales weighting on model performance is demonstrated in Fig.~\ref{fig:sales_weighting_demo}.  Note that given Eqs.~\eqref{eq:reported_metrics}, we want relative metrics to be as close to zero as possible. We find that for region II device C $ {\rm{log\_sales}}~(w_i = log(y_i) ~\rm{in}~ Eq.~\eqref{eq:weighted_mse})$ is the best weight for linear learners and $ {\rm{sales}} ~(w_i = y_i)$ is the best weight for tree learners. We use XGB for the demonstration here as we have seen it's one of the top-performing model for our use case. Note, that behavior is qualitatively similar regardless of the model choice though.  

When we use sales weighting in cost function, we introduce a positive bias in logarithmic units. Due to Jensen's inequality, a positive bias in log units translates to less negative bias in original units. As we progressively, make the weights more aggressive, at some point, introduced bias in log units is just right enough to obtain zero bias in original units. (This is illustrated in Fig.~\ref{fig:sales_weighting_log_demo}.) 

Addition of weights in cost function is easy to implement in most ML libraries and is seen to give improved performance in most cases. However, it suffers from the following limitations: 1) The process is iterative. We don’t know which is the best weight to use apriori. 2) Performance can widely differ depending on the weight chosen 3) On a scientific front, it doesn’t tackle the bias problem head on. This motivates us to explore methods that correct for transformation bias directly.

\subsection{Prediction-based bias correction}
\label{subsec:PBBC}

In the simplest case of explicit bias transformation, the `true' expected value of the back-transformed distribution is related to the back-transformed value obtained from model by a multiplicative bias correction (BC) factor. 
\begin{equation}
\label{eq:BC_intro}
E[Y] = BC * f^{-1} (E[f(Y)])
\end{equation}
To relate Eq.~\eqref{eq:BC_intro} to Eq.~\eqref{eq:JI_back_transformed}, note that $Y=sales$ and $f = log$. Also, based again on Eq.~\eqref{eq:JI_back_transformed}, we know that $BC \geq 1$.  

According to \cite{Forestry_paper, newman}, the bias correction factor is related to the residuals by 
\begin{equation}
BC \equiv e^{\epsilon} = \dfrac{\sum_{i=1}^N e^{\epsilon_i}}{N} \;.
\label{eq:mean_var_correction}
\end{equation} 
On incorporating the bias correction from Eq.~\ref{eq:mean_var_correction}, we find that it helps mitigate the bias (by up to $50\%$), but doesn't eliminate it (Please refer to Fig.~\ref{fig:mean_based_bc} for details). We will refine this bias correction factor next.

Fig.~\ref{fig:in_sample_model_fit} informs us that points with higher sales need a stronger bias correction. However, the BC multiplier from Eq.~\eqref{eq:mean_var_correction} is independent of the sales. This motivates us to make bias correction a function of sales. Obviously, such a correction would have limited applicability, because we don’t know the sales when we are making out of sample predictions. Therefore, we use model prediction, $\hat{Y}$, as a proxy for sales. Revisiting Eq.~\eqref{eq:BC_intro}
\begin{equation}
BC(\hat{Y}) = \dfrac{E[Y]}{E[f^{-1}(\widehat{f(\hat{Y})})]}
\label{eq:pred_based_bc}
\end{equation}
The numerator  $E[Y]$ in Eq.~\eqref{eq:pred_based_bc} is the mean of the original sales data and the denominator in Eq.~\eqref{eq:pred_based_bc} is the mean of uncorrected predicted value. Instead of calculating the BC factor over whole sample (as in Eq.~\eqref{eq:mean_var_correction}), we evaluate the right hand side of Eq.~\eqref{eq:pred_based_bc} over specific prediction windows. Essentially, we learn a piece-wise linear/step function for bias correction.

We divide the prediction region in multiples of 2 (in logarithmic units) and evaluate the right hand side of Eq.~\eqref{eq:pred_based_bc}. 
For the test case of region II device B,  the function we learn for bias correction is shown in Table~\ref{table:bc_pred_based_multiplier}.
\begin{table}
\caption{Prediction-based BC function for region II device B}
\centering
\label{table:bc_pred_based_multiplier}
	\begin{tabular}{| c c c  |}
		\hline
		log\textunderscore vdp\textunderscore plus\textunderscore units (or prediction) &	XGB\textunderscore linear	 & XGB\textunderscore tree 
 \\
		\hline
		$[0-2)$ &	1.503  &	1.015
\\
	$[2-4)$ &	1.142  &	1.093
\\
		$[4-6)$ &	1.493  &	1.099
\\
		$[6-8)$ &	2.112  &	 1.351
\\
		$\geq$ 8 &	2.527  &	1.687
\\		
		\hline
	\end{tabular}
\end{table}
\begin{figure}
\centering
	\includegraphics[width=0.9\linewidth]{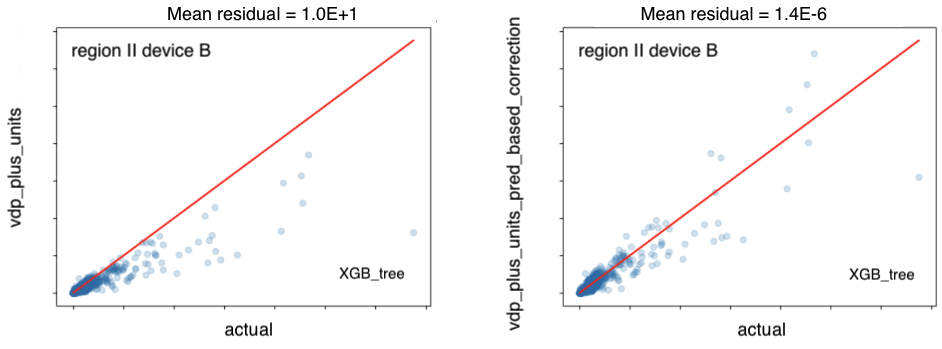}
	\caption{In-sample model fit for XGB model. The plot on the left is obtained by back-transforming the output from ML model without any bias correction. Plot on right is obtained by multiplying the model prediction by prediction-based bias correction (Table~\ref{table:bc_pred_based_multiplier}) }
	\label{fig:PB-BC_in_sample}
\end{figure} 

We see that the prediction based bias correction gets rid of bias in the back-transformed units. This is evident from the mean of residuals being very close to zero in the plot in right panel in Fig.~\ref{fig:PB-BC_in_sample}. In applying the PB-BC technique to out-of-sample data, we make the assumption that the correction function learnt (e.g., in Table~\ref{table:bc_pred_based_multiplier}) carries over to the unseen data as well. There are some obvious improvements that can be done to the prediction-based correction function learnt above. For example, the step width is not uniformly spaced in the original units and the number of data points aren’t evenly distributed in each prediction window.

\begin{figure}
\centering
	\includegraphics[width=0.9\linewidth]{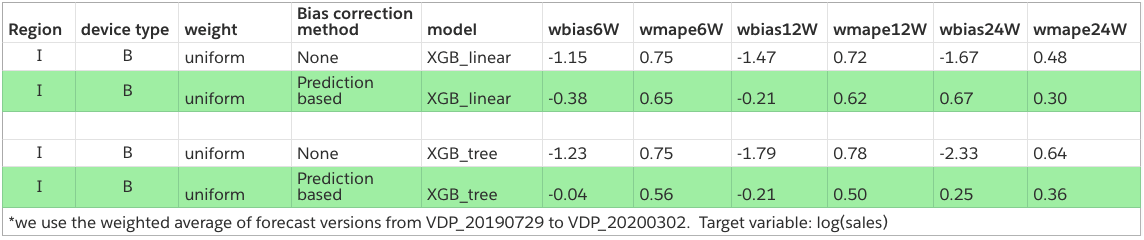}
	\caption{Improvements from prediction-based bias correction for region I device type B.}
	\label{fig:PB-BC_metrics}
\end{figure} 
In Fig.~\ref{fig:PB-BC_metrics} we look at the out-of-the-sample performance in region I device B after incorporating prediction-based bias correction. Takeaways from Fig.~\ref{fig:PB-BC_metrics} are: 1) the prediction-based bias correction immensely helps the performance.  As expected the largest improvement is in the bias metrics.
2) the benefit from bias correction transfers into MAPE improvement as well. The largest MAPE improvement of $56\%$ ($39\%$ → $25\%$) is in 12W MAPE.

We acknowledge that MAPE values from PB-BC may not always perform well especially when compared to addition of weights in cost function. An example of this is show in 
Fig.~\ref{fig:PB-BC_metrics_II}. A possible reason is that the window-based approach used for calculating PB-BC may not be optimal due to data sparsity for some of the prediction windows. Checking alternative approaches of estimating PB-BC function is an ongoing effort. 

\section{Circumventing Transformation Bias} \label{sec:circumvent_tb}

Performance of any bias correction method will depend on how well it is able to estimate the induced transformation bias. An alternate approach to tackle this issue head-on will be to not introduce the transformation bias in the first place by working directly with actual sales as target variable.

\subsection{Introduction to Tweedie Regression}

We established in Sec.~\ref{subsec:model_on_sales} that building a model on sales with MSE as a cost function led to poor performance due to skewed nature of sales data. One of the main ways in which we will generalize our regression approach will be to replace MSE by ‘deviance’ of a distribution in the Tweedie family \cite{Tw1, Tw2, Tw3}.

Tweedie distribution are a family of probability distribution that include Normal, Poisson, and Gamma distributions as special cases. 
This is motivated in Eqs.~\eqref{eq:Tweedie}.  Tweedie distributions are characterized by the 
`tweedie variance power', p. The mathematical formulas for `tweedie deviance’ (analogous to the cost function) are shown in Eq.~\eqref{eq:Tweedie}. 
\begin{eqnarray}
&\bullet ~{\rm{Poisson:}}~ &2 \sum_i \left(y_i log\frac{y_i}{\hat{y}_i} - y_i + \hat{y}_i\right)\nonumber \\
&\bullet ~{\rm{Gamma:}}~ &2 \sum_i  \left( -log\frac{y_i}{\hat{y}_i} + \frac{y_i - \hat{y}_i}{\hat{y}_i} \right) \nonumber \\
&\bullet ~{\rm{Tweedie:}}~ &2 \sum_i \left( y_i \frac{y_i^{1-p} - \hat{y}_i^{1-p}}{1-p} - \frac{y_i^{2-p} - \hat{y}_i^{2-p}}{2-p} \right) \nonumber \\
&\bullet ~{\rm{Normal:}}~  ~~& \sum_i (y_i - \hat{y}_i)^2 
\label{eq:Tweedie} 
\end{eqnarray}
Its straightforward to see that:
\begin{itemize}
\item
setting $p = 0$, recovers MSE;  $p \rightarrow 1$, recovers Poisson; and $p \rightarrow 2$, recovers Gamma
\item
$1 < p < 2$ is referred to as Compound Poisson Gamma or just Tweedie. In rest of the discussion, we us the term Tweedie to refer to this case.  
\end{itemize}

$y_i$ and $\hat{y}_i$ in Eq.~\eqref{eq:Tweedie} are the data points and fitted values respectively. Just as while optimizing for MSE, we assume that the conditional distribution of target variable is normally distributed around the point estimate, while optimizing for Tweedie deviance, we assume that the distribution is Tweedie distributed around the point estimate output from the model. 

Literature applications of Tweedie regression is typically for long-tailed positive data such as number of insurance claims/ rainfall events per year \cite{Tw_insurance}. An advantage of using Tweedie for sales forecasting is that the output is always positively valued. (When using MSE, the output can be negative as well.) 

\subsection{Experimental set up}
\label{subsec:exp_set_up}
We will next look at results with using Tweedie cost function. The goal of these experiments is to test the effect of model design choice (specifically target transformation and cost function) on performance. To achieve this, we choose the same algorithm (XGB), feature set, and hyper-parameters.  

The design choices that we evaluate are the following: 
\begin{enumerate}
\item
Using actual sales as target variable and MSE as the cost function
\item
Using actual sales as target variable and pseudoHuber loss as the cost function. We use pseudoHuberloss as the twice differentiable alternative to mean absolute error (MAE). 
\begin{enumerate}
\item
A motivation for using pseudoHuberloss is that it is closely aligned with MAPE which is our model evaluation metric. 
\item
We also note that optimizing on MAE/ MAPE outputs median sales distribution for a given item for a given day. Medians aren't additive and therefore summing up daily sales estimate to obtain for example estimate for weekly sales is incorrect. This limits the practical usability of models optimized on MAE.  
\end{enumerate}

\item
Using the actual sales as target variable and tweedie deviance as cost function. Experiments were done using the tweedie variance power 1.1, 1.3, 1.5, 1.7, and 1.9. 
\item
Using log(sales) as target variable and not doing any bias correction. We know from Sec.~\ref{sec:root_cause} that this is suboptimal and these results are only for reference.
\item
Using log(sales) as target variable and sqrt(sales) as weight. The weight is added to correct for transformation bias. This is the dominant setting used in production and we use this as a benchmark.  
\end{enumerate}

\subsection{Tweedie regression results}
\label{subsec:tweedie_regression_results}

\begin{figure}
\centering
	\includegraphics[width=0.9\linewidth]{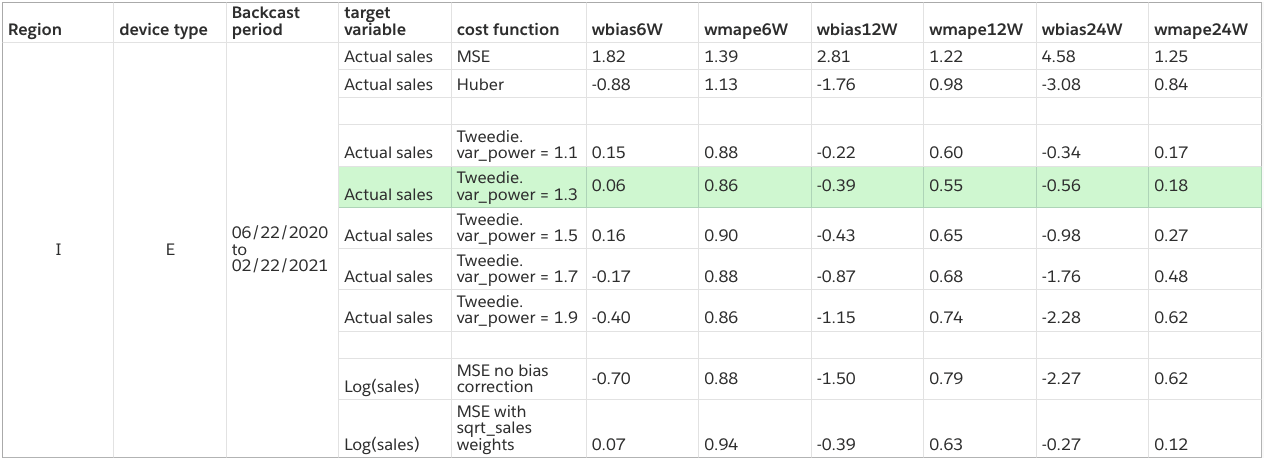}
	\caption{Performance of various Tweedie models for region I device type G}
	\label{fig:tw_I_G}
\end{figure}

In Fig.~\ref{fig:tw_I_G} we present results for a specific backtesting period for one of the region device type combinations. Performance for rest of the combinations is in Appendix~{\ref{appendix:bc_tw_results}}. Also note that in the results presented below, we report the performance of tree-based learner in the XGB algorithm as its almost always better than linear learner. While evaluating MSE and Huber cost function we set the negative predictions if any to zero. 

Observations from the Tweedie regression results are as follows:
\begin{enumerate}
\item
The model built on actual sales with MSE as cost functions performs very poorly consistently. 
\item
The model built using pseduoHuber loss performs much better than MSE, but still has high MAPE/bias values. 
\item
For few of the regions (e.g, I and IV), the Tweedie regression (TR) models perform better than XGB models with bias correction (BC) — current benchmark. For many other countries, TR performance is comparable to the current benchmark.  A detailed survey of TR performance can be found in Appendix~\ref{appendix:bc_tw_results}. 
\item
Variance power of 1.1--1.3 gives the best results for majority of the cases.
\item
Bias values in TR are sensitive to the variance power. Specifically, as the variance power is increased the bias decreases (become more negative if it is around zero at $\rm var\textunderscore power = 1.1$) 
\end{enumerate}

\subsection{Making sense of TR results}

An unmistakable pattern in the Tweedie results (Fig.~\ref{fig:tw_I_G}) is that model using MSE as cost function has a very high bias and hence a high MAPE.  For Tweedie, we see a pattern of decreasing bias (not always a good thing because bias can have high negative values) with increase in the variance power — a hyperparameter in the Tweedie regression.  

To make sense of this, we plot the deviance/ cost function (refer Eqs.~\eqref{eq:Tweedie}) for various values of variance power, p. Specifically, we set actual = 100 (an arbitrary choice) and plot the tweedie deviance as prediction varies between 10 and 190. For reference, we also plot the MSE and MAE cost functions.
\begin{figure}
\centering
	\includegraphics[width=0.6\linewidth]{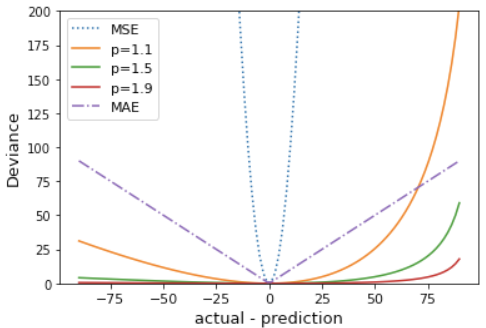}
	\caption{Tweedie deviance for variance power 1.1, 1.5, and 1.9. MSE and MAE are plotted for reference.}
\end{figure}
We see that the deviance for 1.1 is more convex than 1.5 which in turn is more convex than 1.9. This implies that p=1.1 is more likely to give higher prediction values than p=1.9. Compared to these, MSE is much more convex and naturally gives much higher bias. 

On an unrelated note, tweedie deviance functions are asymmetric around zero, penalizing under-prediction more than over-prediction. This might be desirable given that business cost of lost sales is higher than having an overhang.

\section{Conclusion and Next steps} \label{sec:conclusion}

In this work, we demonstrated that a concave transformation of target variable will introduce a negative bias in regression models which is a manifestation of Jensen's inequality. We proposed two methods to correct the transformation bias 1) addition of weights in cost function 2) direct prediction-based bias correction. These bias correction methods significantly improve both MAPE (up to $50\%$) and bias (up to $20$X, e.g., refer Fig.~\ref{fig:PB-BC_metrics})  over different forecasting horizons. 

The optimal method for bias correction is best chosen by experimentation. Sales weighting is easy to implement in most ML libraries but requires more experimentation. On the other hand, PB-BC has a more involved implementation, but involves less experimentation (once the piece-wise linear correction is determined).

We also found that leveraging Tweedie family of cost functions is a great alternative when constructing model directly on sales without any transformation. This allows us to circumvent the issue of transformation bias by avoiding target transformation altogether. We observed that tweedie variance power has a marked effect on the bias metrics and we were able to explain this based on the cost function behavior as a function of variance power. 

In our work, we treated variance power as a hyper-parameter and chose it depending on the values which gave the best backtest metrics. A related question, we ask next is if we can estimate the best tweedie variance power without backtesting. This involves looking at the distribution of \emph{deviance residuals}. Deviance residuals are generalization of residuals when using generalized cost functions (refer Eq.~\eqref{eq:Tweedie}). We expect deviance residuals to be normally distributed if the model chosen closely reflects the underlying data generation process. This forms the basis of our ongoing work.

\section{Science and Business Outlook}

Even though we exclusively focused on forecasting models in this paper, the issues and methods discussed are relevant anytime we have a right-skewed distribution. Log transformation is a standard practice in many regression settings and this work highlights the risk associated with it. Moreover, the methods presented in this paper such as changing the cost function to a weighted MSE or a Tweedie deviance are relatively straightforward and easy to implement in most business settings.  

\section*{Acknowledgments}

Author would like to thank Amazon colleagues -- Dhruv Madeka,   Roberto Ayala,  Dinesh Mandalapu,  Farhad Ghassemi,  Zhexiang Sheng,  Longshaokan (Marshall) Wang,  Oinam Nganba Meetei,  and De Chen for insightful discussions and valuable feedback.

\input{appendix}

\end{document}

%% file: appendix.tex
\begin{appendices}
	\appendix

\section{VDP model metrics} \label{sec:model_metrics}

Performance of various VDP models built on transformed target variable is shown in Fig.~\ref{fig:various_ models_neg_bias}. The performance is measured across different horizons (6 week, 12 week, and 24 week). We generate a new forecast every week labeled by VDP\textunderscore YYYYMMDD. The MAPE (mean absolute percentage error) and bias metrics are generated for each forecast version. The reported metrics for a given model are sales-weighted average over multiple forecast versions. 

The details of the WMAPE (weighted MAPE) and wbias (weighted bias) calculation are as follows:
\begin{flalign}
	& F_{i, V, d}: ~{\rm Forecasted ~units~ for~  item~ \emph{i}~ for~ forecast~ version~ \emph{V}~ on~  day~ \emph{d}} \nonumber \\ 
	& A_{i, V, d}: ~{\rm Actual ~units~ for~  item~ \emph{i}~ for~ forecast~ version~ \emph{V}~ on~  day~ \emph{d}} \nonumber \\
	& A_{i,V}:~{\rm Actual~units~for~item~\emph{i}~for~forecast~version~\emph{V}~across~time~horizon} \nonumber \\
	& h:~{\rm set~of~all~days~in~time~horizon~of~metric} \nonumber \\
	& PE_{i,V} = \dfrac{\sum_{i \in h} F_{i,V,d} - \sum_{i \in h} A_{i,V,d}}{\sum_{i \in h} A_{i,V,d}} \nonumber \\
	& WMAPE_{V} = \dfrac{\sum_i A_{i,V} * \left| PE_{i,V} \right|}{\sum_i A_{i,V}} ~\;\;\;~ WBias_{V} = \dfrac{\sum_i A_{i,V} * PE_{i,V}}{\sum_i A_{i,V}}
	\label{eq:mape_bias_def}
\end{flalign}

%

%

\section{Bias correction details}



\subsection{Sales weighting effect}

\begin{figure}
\centering
	\includegraphics[width=0.9\linewidth]{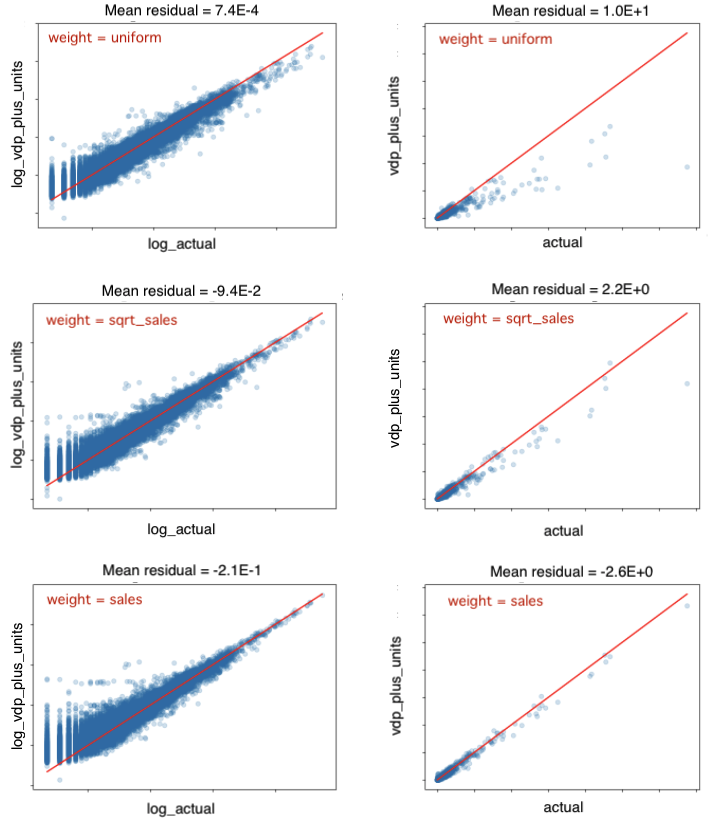}
	\caption{In-sample model fit for region II device B for XGB\_tree using different weights in the cost function. The plots on the left are the fit in the modeling units (logarithm in this case). The Y axis is the prediction output from the ML model. The plots on the right are obtained by back-transforming (exponentiating in this case) the values from plots on the left. In particular, we don’t do any model refit while going from plot on left to the plots on right.  }
	\label{fig:sales_weighting_log_demo}
\end{figure}

When we use sales weighting in cost function, we introduce a positive bias in logarithmic units. Due to Jensen's inequality, a positive bias in log units translates to less negative bias in original units. As we progressively, make the weights more aggressive, at some point, introduced bias  in log units is just right enough to obtain zero bias in original units. This is illustrated in Fig.~\ref{fig:sales_weighting_log_demo}.  
Note that $\rm residual = actual - prediction$.  So, negative bias $\Rightarrow \rm  mean(residual) > 0$

\subsection{Direct bias correction}

According to \cite{Forestry_paper, newman}, the bias correction factor is related to the variance of residuals as 
\begin{equation}
BC = e^{\sigma^2 / 2}\;.
\label{eq:variance_correction}
\end{equation}
Eq.~\eqref{eq:variance_correction} holds when the residuals are normally distributed. If residuals are not normally distributed, `a smearing estimate of bias' is recommended:
\begin{equation}
BC \equiv e^{\epsilon} = \dfrac{\sum_{i=1}^N e^{\epsilon_i}}{N}
\label{eq:mean_correction}
\end{equation} 
where $\epsilon_i$ is the $i^{\rm th}$  regression residual. We calculate BC multiplier from Eqs.~\eqref{eq:variance_correction} and \eqref{eq:mean_correction} for test case of region II device B. The multipliers obtained are shown in Table~\ref{table:bc_multipliers}.

\begin{table}
	\caption{Bias correction multipliers for region II device B}
	\centering
	\label{table:bc_multipliers}
	\begin{tabular}{| c c c  |}
		\hline
		 &	XGB\textunderscore linear	 & XGB\textunderscore tree 
 \\
		\hline
		Variance-based multiplier [Eq.~\eqref{eq:variance_correction}]  &	1.29  &	1.07	
\\
		Mean-based multiplier [Eq.~\eqref{eq:mean_correction}] 
			& 1.27 & 1.07
\\
		
		\hline
	\end{tabular}
\end{table}

BC multipliers from Eq.~\eqref{eq:variance_correction} and \eqref{eq:mean_correction} are similar as shown in Table~\ref{table:bc_multipliers}  because for these models the residuals are normally distributed to a good approximation.

\begin{figure}
\centering
	\includegraphics[width=\linewidth]{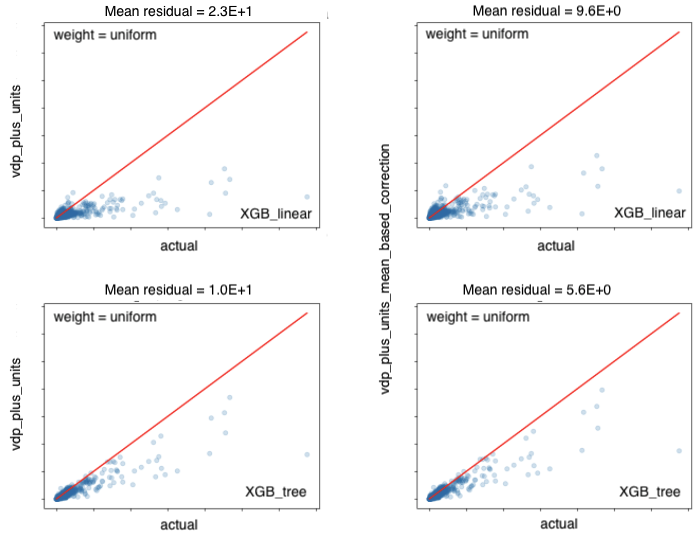}
	\caption{In-sample model fit for region II device B for XGB model. The plots on the left are obtained by back-transforming the output from ML model without any bias correction. Plots on right are obtained by multiplying the model prediction by mean-based bias correction (Eq.~\eqref{eq:mean_correction}). }
	\label{fig:mean_based_bc}
\end{figure}

In Fig.~\ref{fig:mean_based_bc}, we plot the raw uncorrected back-transformed values on left (these are similar to plots on right in Fig.~\ref{fig:in_sample_model_fit}). The plots on right in Fig.~\ref{fig:mean_based_bc} are obtained by correcting for bias transformation using Eq.~\eqref{eq:mean_correction}. Specifically, we multiply the model predictions by second row in Table~\ref{table:bc_multipliers}. 
Few observations, we make from Fig.~\ref{fig:mean_based_bc}:
\begin{enumerate}
\item
The negative bias definitely decreases after correction. This can be seen from the mean of residuals moving closer to zero. In fact, for both XGB\textunderscore tree and XGB\textunderscore linear, the bias decreased by about $50\%$.
\item
Nonetheless, we are still far from completely removing bias. For instance, the mean of residuals in log units (left panel in Fig.~\ref{fig:in_sample_model_fit}) is much closer to zero than what we have in the original units even after bias correction
\item
We notice that points with higher sales show a more prominent bias than lower sales. This motivated us to derive a ‘prediction-based’ bias correction.
\end{enumerate}

\subsection{Bias correction derivation}

Here we derive the BC factor in Eq.~\eqref{eq:variance_correction}. Following the approach in \cite{newman}, in case of linear regression we have:
\begin{equation}
log(Y) = b_0 + b_1 X + \epsilon
\end{equation}  
$\epsilon$ has a mean 0 in logarithmic units but not in the original arithmetic units.  Because the mean will not be zero after back-transformation, the error term must be retained during back transformation:
\begin{equation}
Y = b_0 e^{b_1 X} e^{\epsilon}
\label{eq:exp_epsilon}
\end{equation}

$e^\epsilon$ is the bias correction factor (Eq.~\eqref{eq:mean_correction}). If we assume that $\epsilon$ follows a normal distribution with mean $0$ and variance $\sigma^2$, 
\begin{equation}
E[e^\epsilon]  =  \int_{-\infty}^{\infty} e^{\epsilon} \mathcal{N}(0; \sigma^2) \,d\epsilon 
                        =  e^{\sigma^2/2}
\label{eq:bc_variance_derivation}
\end{equation}
Eq.~\eqref{eq:bc_variance_derivation} is the bias correction factor we had in Eq.~\eqref{eq:variance_correction}. 

Another point to note is that 
\begin{eqnarray}
E[e^{\epsilon}] & \geq & e^{E(\epsilon)} \\
\Rightarrow E[e^{\epsilon}] & \geq & 1 \;.
\label{eq:BC_greater_than_1}
\end{eqnarray}
In Eq.~\eqref{eq:BC_greater_than_1}, we use the property that $E(\epsilon) = 0$ or that mean of residuals is zero in the modeling units. Note that we don't make any assumptions on the distribution of error term, $\epsilon$, in the derivation of Eq.~\eqref{eq:BC_greater_than_1}. 

Equation~\eqref{eq:BC_greater_than_1} tells us that the Bias correction factor will always be greater than equal to 1. Implying that our back-transformed forecasts will always under-forecast on average. 

\section{Bias correction and Tweedie Results}
\label{appendix:bc_tw_results}

\begin{figure}
\centering
	\includegraphics[width=0.9\linewidth]{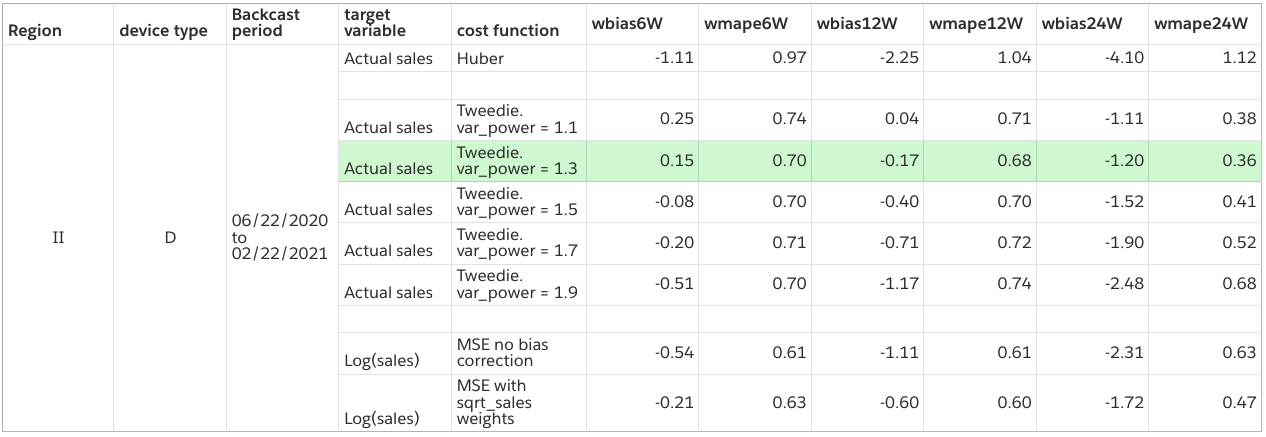}
	\caption{Performance of various Tweedie models for region I device type B}
	\label{fig:I_B_tw}
\end{figure}

\begin{figure}
\centering
	\includegraphics[width=0.9\linewidth]{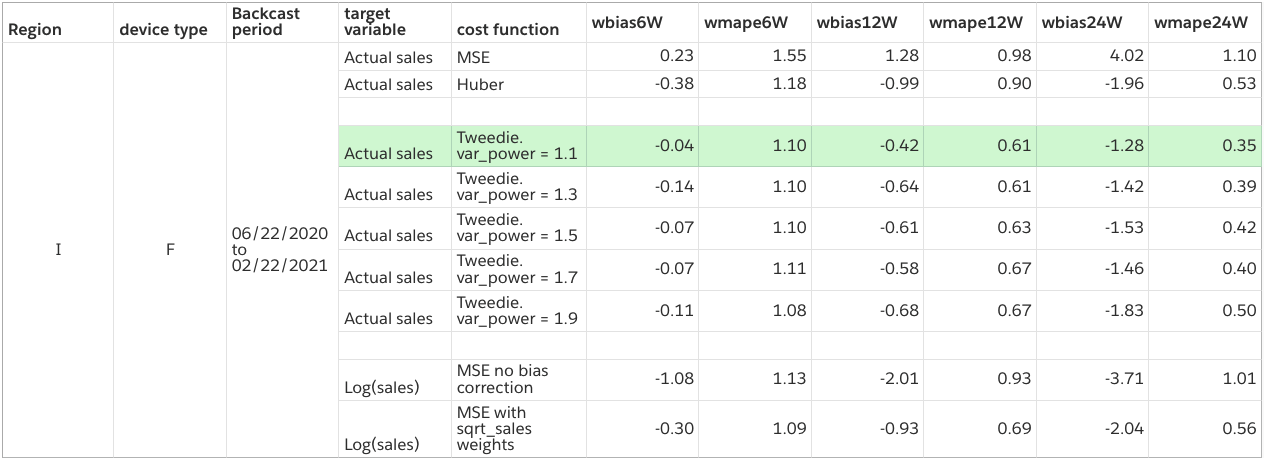}
	\caption{Performance of various Tweedie models for region I device type F}
	\label{fig:I_F_tw}
\end{figure}

\begin{figure}
\centering
	\includegraphics[width=0.9\linewidth]{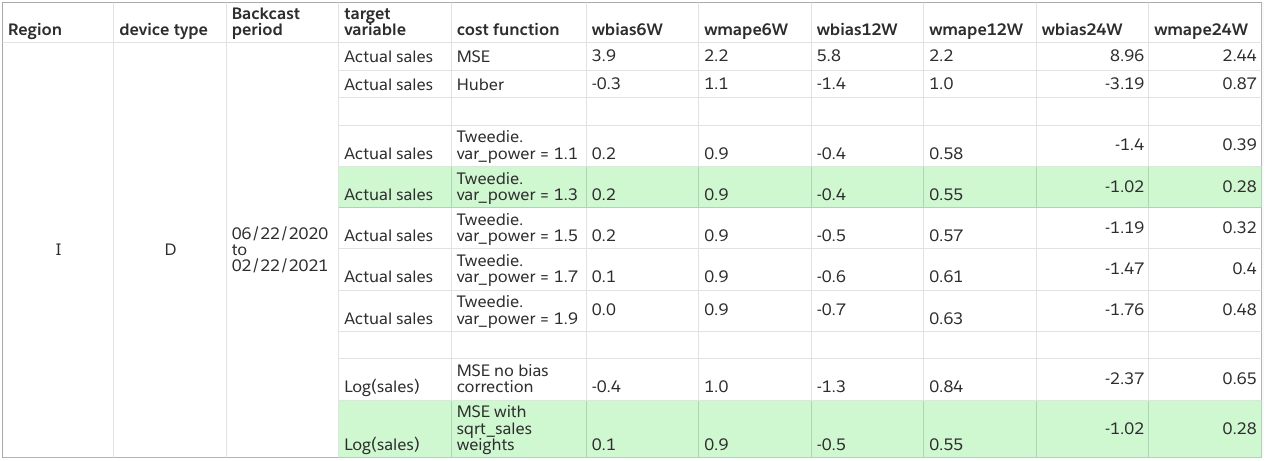}
	\caption{Performance of various Tweedie models for region II device type D}
	\label{fig:II_D_tw}
\end{figure}

\begin{figure}
\centering
	\includegraphics[width=0.9\linewidth]{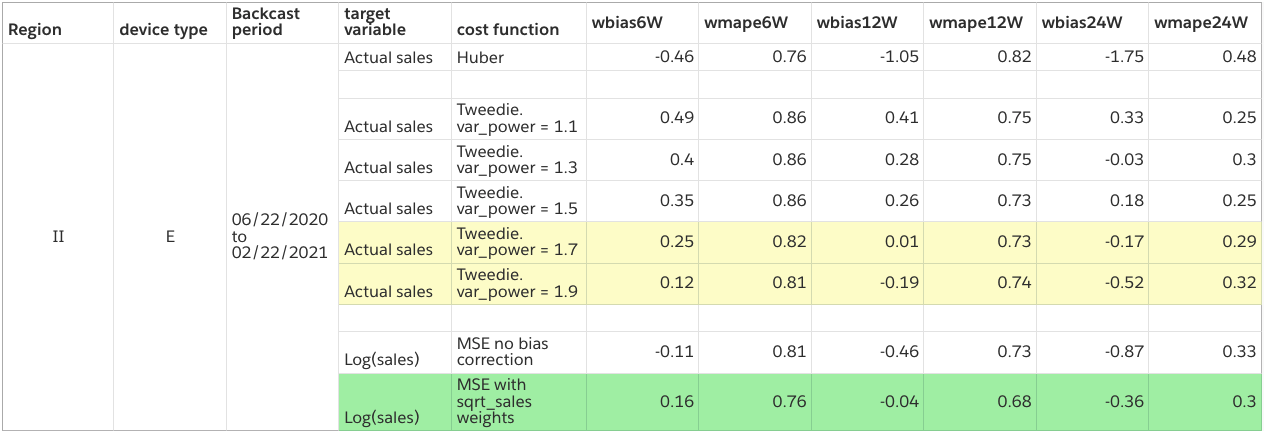}
	\caption{Performance of various Tweedie models for region II device type E}
	\label{fig:II_E_tw}
\end{figure}

\begin{figure}
\centering
	\includegraphics[width=0.9\linewidth]{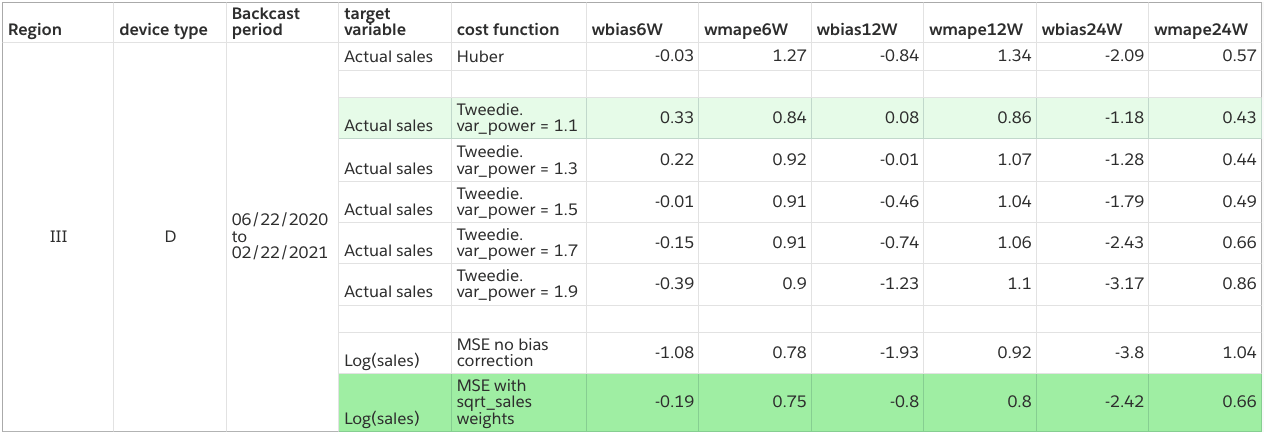}
	\caption{Performance of various Tweedie models for region III device type D}
	\label{fig:III_D_tw}
\end{figure}

\begin{figure}
\centering
	\includegraphics[width=0.9\linewidth]{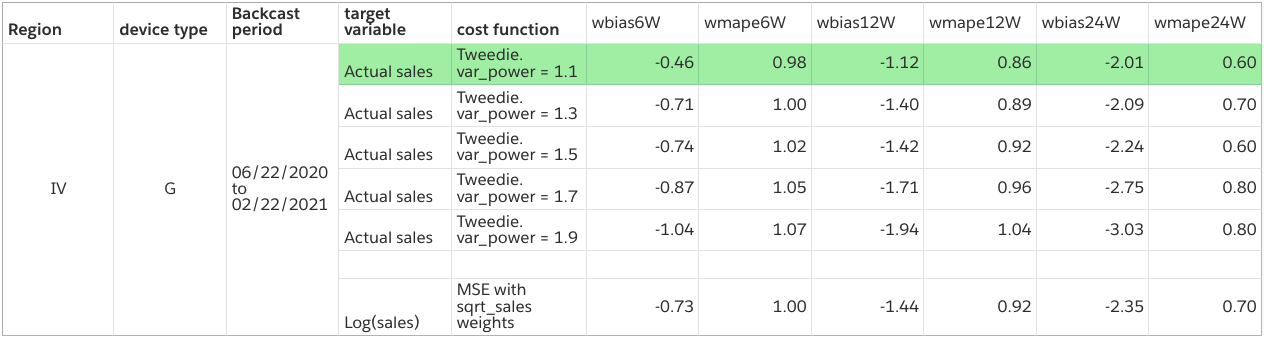}
	\caption{Performance of various Tweedie models for region IV device type G}
	\label{fig:IV_G_tw}
\end{figure}

We provide a survey of Tweedie regression performance in Fig.~\ref{fig:I_B_tw} through Fig.~\ref{fig:IV_G_tw}.  The takeaways from these results are presented in Sec.~\ref{subsec:tweedie_regression_results}.

\begin{figure}
\centering
	\includegraphics[width=0.9\linewidth]{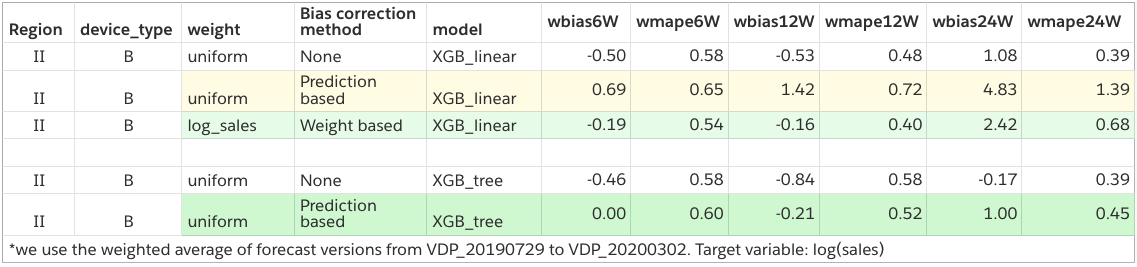}
	\caption{Improvements from prediction-based bias correction for region II device type B}
	\label{fig:PB-BC_metrics_II}
\end{figure} 

%

\end{appendices}